\documentclass[conference]{IEEEtran}

\usepackage{cite}
\usepackage{amsmath,amssymb,amsfonts}
\usepackage{graphicx}
\usepackage{textcomp}
\usepackage{url}
\usepackage{siunitx} 
\usepackage[table]{xcolor} 

\usepackage{algorithm}
\usepackage{algpseudocode} 

\usepackage{booktabs}    
\usepackage{multirow}    
\usepackage{tabularx}    
\usepackage{threeparttable} 
\usepackage{arydshln}

\usepackage{balance}     
\usepackage{soul}        

\definecolor{Accent}{RGB}{0,92,184}
\definecolor{AccentLite}{RGB}{235,245,255}
\definecolor{RowGray}{gray}{0.97}

\definecolor{BgGreen}{RGB}{245,252,245}
\definecolor{BgBlue}{RGB}{245,248,255}
\definecolor{BgOrange}{RGB}{255,250,242}
\definecolor{BgOurs}{RGB}{255, 252, 235}

\newcolumntype{L}[1]{>{\raggedright\arraybackslash}p{#1}}
\newcolumntype{R}[1]{>{\raggedleft\arraybackslash}p{#1}}
\newcolumntype{Y}{>{\raggedright\arraybackslash}X}

\def\BibTeX{{\rm B\kern-.05em{\sc i\kern-.025em b}\kern-.08em
    T\kern-.1667em\lower.7ex\hbox{E}\kern-.125emX}}
    
\newif\ifanonymous
\anonymousfalse 
\IEEEoverridecommandlockouts
\begin{document}

\title{LSTrans: Efficient Knowledge Transfer for Lightweight and Automated ECG Classification}

\ifanonymous
  \author{\IEEEauthorblockN{Anonymous Author(s)}
  }
\else
\author{\IEEEauthorblockN{Yi Zhao}
\IEEEauthorblockA{School of Mechano-Electronic Engineering \\
Xidian University\\
Xi'an, China \\
24049200262@stu.xidian.edu.cn}
\and
\IEEEauthorblockN{Jiajun Gao}
\IEEEauthorblockA{School of Economics and Management \\
Xidian University\\
Xi'an, China \\
25069100156@stu.xidian.edu.cn}
\and
\IEEEauthorblockN{Chenyang Xu}
\IEEEauthorblockA{School of Cyber Engineering \\
Xidian University\\
Xi'an, China \\
xcy@ieee.org}
\and
\IEEEauthorblockN{4\textsuperscript{th} Yuxi Zhou\textsuperscript{*}}
\IEEEauthorblockA{\textit{DCST, BNRist, RIIT, Institute of Internet Industry} \\
\textit{Tsinghua University}\\
Beijing, China \\
joy\_yuxi@pku.edu.cn\thanks{*Corresponding authors: Yuxi Zhou (joy\_yuxi@pku.edu.cn) and Hao Wang (Haow@ieee.org).}}
\and
\IEEEauthorblockN{5\textsuperscript{th} Hao Wang\textsuperscript{*}}
\IEEEauthorblockA{\textit{School of Cyber Engineering} \\
\textit{Xidian University}\\
Xi'an, China \\
Haow@ieee.org}
}
\fi

\maketitle

\begin{abstract}
Deploying deep learning models for automated electrocardiogram classification on resource-constrained wearable devices remains challenging due to high computational costs. To address this, we propose LSTrans, a lightweight hybrid model designed for efficient and sensitive ECG analysis. 
LSTrans introduces a specialized 1D convolutional backbone with an interleaved layer architecture to capture both macroscopic rhythmic trends and microscopic morphological variations. This backbone is cascaded with a Transformer encoder to model long-range temporal dependencies, incorporating Low-Rank Adaptation across critical layers to compress the model and reduce the trainable parameter space.
We further employ homogeneous and heterogeneous knowledge distillation to transfer diagnostic expertise from high-capacity teacher models to the student. 
Experimental results on multiple benchmark datasets demonstrate that LSTrans achieves a competitive balance between diagnostic sensitivity and resource efficiency, substantially reducing peak memory footprints and training latency during downstream adaptation. 
The source code is available for review at \url{https://github.com/zyee00128/LSTrans4BIBM}.
\end{abstract}


\section{Introduction}
\label{sec:intro}
Cardiovascular diseases (CVDs) remain a major cause of global mortality. This issue drives the demand for real-time cardiac monitoring. Electrocardiogram (ECG) analysis is the standard clinical tool to identify heart abnormalities. Deep neural networks have achieved cardiologist-level performance in ambulatory ECG classification\cite{hannun2019cardiologist}. They also demonstrate high accuracy in diagnosing pathologies across standard 12-lead systems\cite{ribeiro2020automatic}. Modern consumer-grade wearables now allow continuous, remote ECG screening outside of hospitals\cite{wang2024systematic,tegegne2026wearable}. However, deploying these deep learning models directly on edge devices or Internet of Healthcare Things (IoHT) nodes is difficult. Such resource-constrained hardware units enforce strict memory boundaries and low-power limits, which restrict the deployment of standard, heavy deep networks\cite{manimaran2025explainable,guhdar2025advanced}.

Healthcare foundation models have shown the value of scaling representations for ECG tasks\cite{han2024foundation}. However, running or fine-tuning high-capacity networks on wearable sensors requires parameter-efficient adaptations to save memory\cite{nandakishor2025high,zhou2025h}.
Accurate ECG analysis must process macroscopic rhythm trends alongside microscopic localized waves\cite{narotamo2024deep,strodthoff2020deep}. Advanced architectures address this through complex designs. For example, ECGTransForm\cite{el2024ecgtransform} utilizes multi-scale convolutions with bidirectional Transformers to capture spatial-temporal features. Similarly, MTA-Net\cite{pang2025mta} adopts DWPT for fine-grained time-frequency representations. Nevertheless, their multi-branch configurations often incur heavy computational overhead\cite{bai2018empirical}. Lightweight networks like CrossStateECG-Lite\cite{zheng2025crossstateecg} address resource constraints by employing multi-scale fusion with efficient attention. Yet, these methods are typically limited to single-lead biometric settings rather than multi-lead, multi-label clinical diagnoses.
To bypass this, some methods convert 1D ECG signals into 2D spectrograms or images. For instance, HMT-KD\cite{beigzadeh2026hmt} implements a hierarchical multi-teacher distillation on 2D ECG image representations to train a compact student. However, this 2D conversion can cause phase alignment errors and computational overhead\cite{narotamo2024deep,shivashankara2024ecg}. Simple model compression directly reduces network size but often degrades diagnostic recall. This compromise leads to critical false-negative diagnoses, particularly on subtle morphological anomalies.

To address these challenges, we propose LSTrans, a lightweight hybrid model for ECG classification. LSTrans features a specialized 1D interleaved backbone cascaded with a Transformer encoder, optimized via Low-Rank Adaptation (LoRA)\cite{hu2021lora} and structured knowledge distillation\cite{hinton2015distilling,sun2024diredi}. This hybrid design balances high diagnostic recall with minimal downstream tuning latency and memory footprints. Our experiments on multiple benchmark datasets demonstrate the competitive diagnostic performance of our framework.

\begin{figure}[H]
\centering
\includegraphics[width=0.5\textwidth]{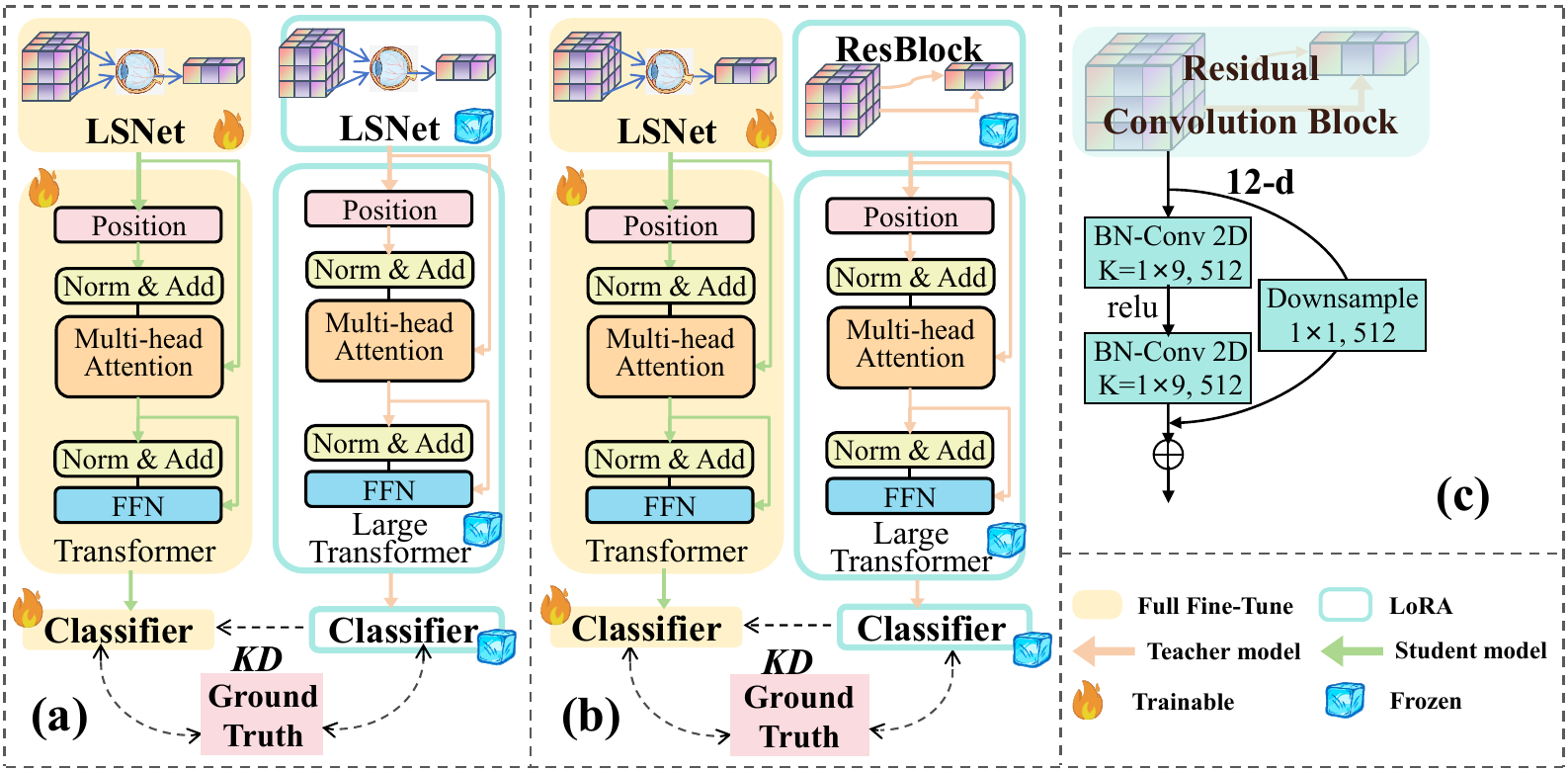} 
\caption{Overview of the knowledge distillation pipeline. (a) Homogeneous distillation with an LSNet teacher. (b) Heterogeneous distillation with a ResNet teacher. (c) The teacher's residual encoder.}
\label{fig:KD Config}
\end{figure}

The key contributions of this work are summarized as follows:

\textbf{Refactored 1D-LSNet for ECG Analysis.} We refactor the 2D heteroscale paradigm into a specialized 1D backbone to process raw ECG signals directly, maximizing computational efficiency while retaining critical morphological representations.

\textbf{Interleaved Architecture for Feature Synergy.} We implement an alternating, interleaved block architecture to balance dual-scale feature capture. This design couples dynamic large-kernel filters for global rhythmic context with multi-scale, lead-attentive units for local morphology, preventing signal degradation during downsampling and minimizing false-negative diagnostic errors.

\textbf{Exploration of Diverse Distillation Strategies.} We explore both homogeneous and heterogeneous knowledge distillation frameworks to bridge the representation gap for our lightweight student. Our systematic analysis shows that structured teacher-student transfer effectively restores diagnostic sensitivity under constrained adaptation resources.

\section{Methodology}
\label{sec: method}

\subsection{Problem Formulation}
\label{subsec: problem formulation}
We frame this task as a multi-label classification problem. Given a fine-tuning dataset $\mathcal{D} = \{(\mathbf{x}_i, \mathbf{y}_i)\}_{i=1}^N$, the vector $\mathbf{y}_i \in \{0,1\}^K$ represents the ground-truth labels for $K$ distinct cardiac abnormalities. We optimize the student model $f_{\theta_s}$ using a combination of these ground-truth labels and knowledge distilled from a pretrained teacher model $f_{\theta_t}$.

Each ECG recording is processed into a tensor $\mathbf{X} \in \mathbb{R}^{C \times L}$, where $C$ represents the number of channels and $L$ represents the temporal length. Section \ref{subsec: datasets} provides the specific preprocessing details. During the training phase, the input data are organized into batches of size $B$. The total loss function is formulated as follows:
\begin{equation}
\label{eq:total_loss}
\begin{split}
\mathcal{L} = \frac{1}{B} \sum_{i=1}^{B} \Big[ &(1 - \alpha) \cdot \ell(\sigma (f_{\theta_s}(\mathbf{x}_i)), \mathbf{y}_i) \\
&+ \alpha \cdot T^2 \cdot \ell \left( \sigma\left(\frac{f_{\theta_s}(\mathbf{x}_i)}{T}\right), \sigma\left(\frac{f_{\theta_t}(\mathbf{x}_i)}{T}\right) \right) \Big],
\end{split}
\end{equation}
where $\ell$ denotes the binary cross-entropy loss, and $\sigma$ represents the sigmoid activation function. The hyperparameter $\alpha \in [0,1]$ balances the hard-label supervision with distillation loss. The temperature factor $T > 0$ scales the logits to smooth the output distribution of the teacher model, which provides more informative soft labels for training the student model.

\subsection{1D-LSNet Backbone}
\label{subsec: LSNet}

\begin{figure*}
\centering
\includegraphics[width=1\textwidth]{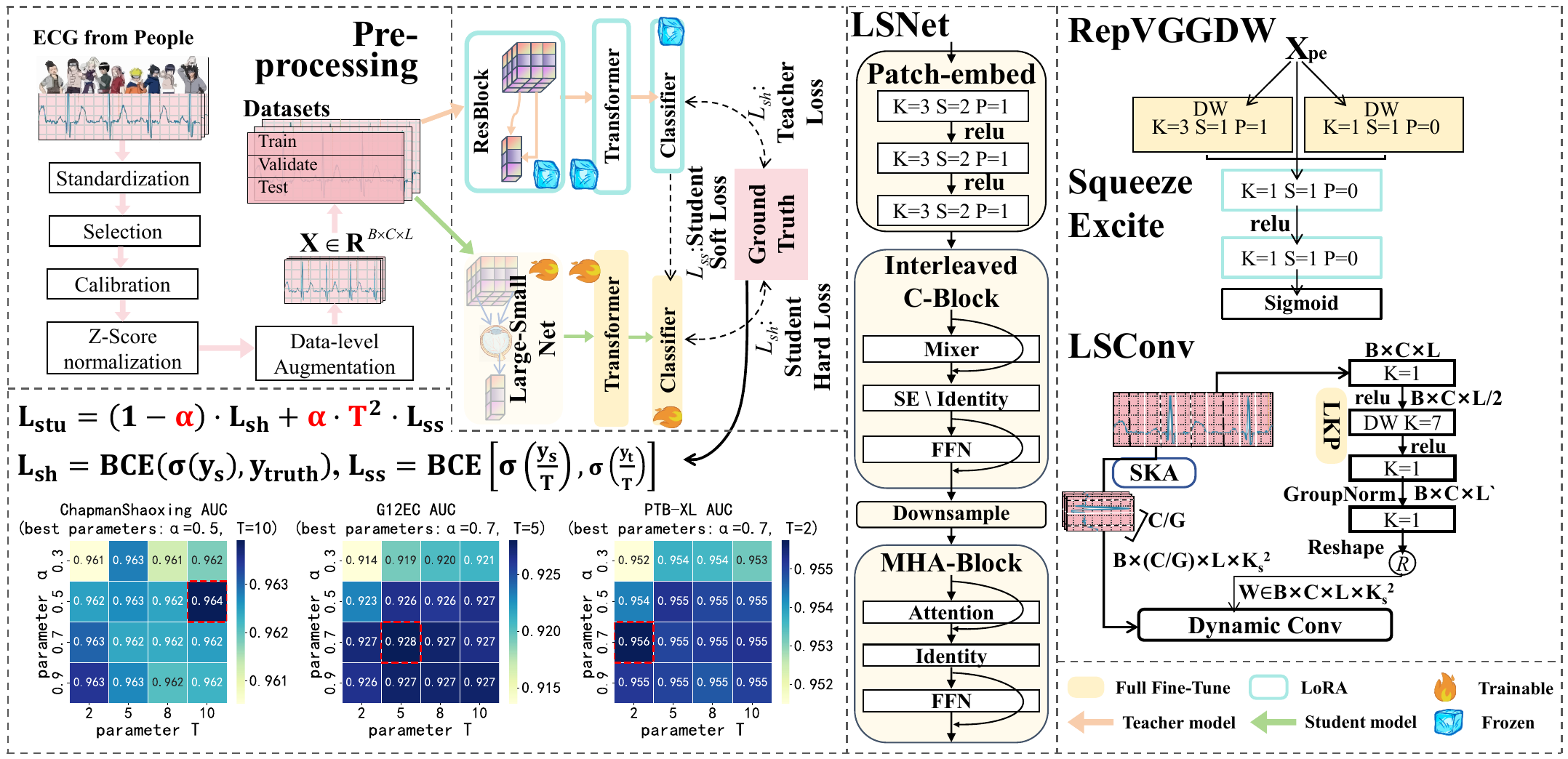} 
\caption{The diagram of the LSTrans heterogeneous framework contains the preprocessing and joint distillation training pipeline, hyperparameter sensitivity heatmaps, and the 1D-LSNet backbone. The backbone includes patch embedding, interleaved blocks, downsampling, and self-attention layers. Within these interleaved blocks, the Mixer component is composed of either the RepVGGDW module or the LSConv block, and SE denotes Squeeze-and-Excite.}
\label{fig:LSTrans Framework}
\end{figure*}

Taking the mini-batch tensor $\mathbf{X} \in \mathbb{R}^{B \times C \times L}$, our 1D adaptation processes raw physiological signals  through hierarchical temporal abstractions. The backbone architecture is designed to reflect the bi-modal clinical characteristics of ECG waveforms. Specifically, it targets localized morphological variations to detect micro-anomalies and extracts global temporal patterns to evaluate macro-rhythms.

\subsubsection{Hierarchical Patch Embedding}
The initial Patch Embedding layer performs 1D feature projection through three successive convolution and batch normalization units, denoted as BN-Conv. The backbone utilizes four stages with depths $d \in \left[1, 2, 8, 10 \right]$. The base projection operation is defined as:
\begin{equation}
    \text{BN-Conv}(\mathbf{X}) = \phi(\text{BN}(\text{Conv}(\mathbf{X})))\ .
\end{equation}
Applying this recursively, the patch embedding output is computed as:
\begin{equation}
    \mathbf{X}_{pe} = \text{BN}(\text{Conv}(\text{BN-Conv}(\text{BN-Conv}(\mathbf{X}))))\in \mathbb{R}^{B \times C_1 \times L_1},
\end{equation}
where $\phi$ denotes the rectified linear unit activation function and $L_i=L/s_i$ represents the reduced temporal dimension at stage $i \in \{0,1,2,3\}$, with $s_i$ denoting the cumulative stride reduction factor. This direct projection avoids the spatial distortions and high-frequency noise that typically arise when transforming sequential physiological leads into two-dimensional matrices.

To transition between feature extraction stages, we avoid non-learnable pooling operations that discard phase information and instead utilize a progressive downsampling strategy driven by strided convolutions. The depthwise convolution block is defined as:
\begin{equation}
    \text{BN-DWConv}(\mathbf{X}) = \text{BN}(\text{DWConv}(\mathbf{X}))\ .
\end{equation}
The downsampled feature map $\mathbf{X}_{ds}$ is then generated by:
\begin{equation}
\label{eq:downsample}
    \mathbf{X}_{ds} = \text{BN-Conv}_{1 \times 1}(\text{BN-DWConv}(\mathbf{X})) \in \mathbb{R}^{B \times C_{i+1} \times L_{i+1}}\ .
\end{equation}
Here, the depthwise convolution incorporates a $1\times3$ kernel. This formulation preserves granular morphological details during spatial reduction, which is critical for retaining subtle structural anomalies.

\subsubsection{Physiological Interleaved Convolutional Architecture}
To capture complementary diagnostic features, we design an alternating, interleaved block architecture. For a block at depth $i$, the output $\mathbf{X}_i$ is computed as follows:
\begin{equation}
\label{eq:ffn}
\mathbf{X}_{mid} = \mathcal{F}_{se}(\mathcal{F}_{mixer}(\mathbf{X}_{i-1})),
\end{equation}
\begin{equation}
\mathbf{X}_{i} = \mathcal{F}_{ffn}(\mathbf{X}_{mid}) + \mathbf{X}_{mid} .
\end{equation}
\begin{equation}
    \mathcal{F}_{ffn}(\mathbf{X}) = \text{BN}(\text{Conv}(\text{BN-Conv}(\mathbf{X}))),
\end{equation}
where the combined operation of the channel-attention layer $\mathcal{F}_{se}$ and the mixer $\mathcal{F}_{mixer}$ is defined as:
\begin{equation}
\resizebox{0.90\linewidth}{!}{%
$\mathcal{F}_{se}(\mathcal{F}_{mixer}(\mathbf{X})) = 
\begin{cases}
\mathcal{F}_{SE}(\mathcal{F}_{rep}(\mathbf{X})), & \text{if } i \text{ is even} \\
\mathcal{F}_{Id}(\mathcal{F}_{ls}(\mathbf{X}) + \mathbf{X}), & \text{if } i \text{ is odd, stage} < 3 \\
\mathcal{F}_{Id}(\mathcal{F}_{attn}(\mathbf{X}) + \mathbf{X}), & \text{if } i \text{ is odd, stage} = 3
\end{cases}$%
}
\end{equation}
Here, $\mathcal{F}_{SE}$ represents Squeeze-and-Excitation, $\mathcal{F}_{rep}$ is the reparameterized module, $\mathcal{F}_{ls}$ is the Large-Small Convolution (LSConv), $\mathcal{F}_{attn}$ denotes the 1D multi-head self-attention mechanism, and $\mathcal{F}_{Id}$ acts as the identity mapping.

This interleaved architecture alternates between even-indexed and odd-indexed layers to capture complementary local micro-morphologies and macroscopic rhythm contexts.

Even-indexed layers focus on localized micro-morphological structures by combining $\mathcal{F}_{rep}$ with $\mathcal{F}_{SE}$. 
The module $\mathcal{F}_{rep}$ functions as a multi-scale wave-front detector, showing high sensitivity to abrupt, localized signal variations such as ST-segment elevations and pathological Q-waves. 
Simultaneously, $\mathcal{F}_{SE}$ block dynamically calculates channel importance weights to address lead-specific spatial distributions, which suppresses background artifacts and highlights diagnostically relevant channels.

In the odd-indexed layers, the channel attention layer $\mathcal{F}_{se}$ is simplified to $\mathcal{F}_{Id}$ to prevent overparameterization and reduce computational redundancy.
The $\mathcal{F}_{mixer}$ is realized as $\mathcal{F}_{ls}$ to filter out baseline noise and capture broad contextual trends. 
This module decouples signal processing into a static $1 \times 7$ depthwise convolution for broad contextual perception and a dynamic block for local feature modeling. 

The transition from localized convolutional feature extraction to global context modeling is achieved at the final stage of the network. We introduce $\mathcal{F}_{attn}$ exclusively in the odd-indexed layers of stage 3. 
At this depth, cumulative downsampling reduces the temporal sequence length to $L_{final} = L/64$. This resolution compresses local wave structures into representative temporal tokens, allowing $\mathcal{F}_{attn}$ to model long-range diagnostic dependencies such as irregular intervals between distant cardiac complexes. The final representation of the backbone feature is denoted as $\mathbf{X}_{lsnet} \in \mathbb{R}^{B \times C_{final} \times L_{final}}$, where $C_{final}$ represents the dimension of the target output channel.

\subsubsection{RepVGGDW and Squeeze-and-Excitation}
\label{sssec:Rep&SE}
The RepVGGDW module consists of three parallel branches, including a $1 \times 3$ depthwise convolution, a $1 \times 1$ depthwise convolution, and an identity mapping. These branches capture multi-scale temporal patterns:
\begin{equation}
\label{eq: RepVGGDW}
\begin{split}
\mathbf{X}_{rep} = \,&\text{BN-DWConv}_{1\times 1}(\mathbf{X}) \\
&+ \text{BN-DWConv}_{1\times 3}(\mathbf{X}) + \mathbf{X} \in \mathbb{R}^{B\times C_i\times L_i} .
\end{split}
\end{equation}
These parallel branches capture multi-scale patterns during training and are mathematically folded into a single $1\times 3$ kernel during inference to ensure zero latency overhead.

The Squeeze-and-Excitation module $\mathcal{F}_{SE}$ performs adaptive channel-wise recalibration on $\mathbf{X}_{rep}$. It utilizes 1D global average pooling to compress the temporal dimension into a global channel descriptor. A bottleneck network with two $1\times1$ convolutions models the inter-channel dependencies to produce a gating weight vector. Finally, the original features are scaled via broadcasting:
\begin{equation}
\label{eq: SE1}
    \mathbf{X}_{mean} = \frac{1}{L_i} \sum_{j=1}^{L_i} \mathbf{X}_{rep}^{(:, :, j)},
\end{equation}
\begin{equation}
\label{eq: SE2}
    \mathbf{W}_{se} = \sigma(\text{Conv}_{1 \times 1}(\phi(\text{Conv}_{1\times1}(\mathbf{X}_{mean})))),
\end{equation}
\begin{equation}
\label{eq: SE3}
    \mathbf{X}_{se} = \mathbf{X}_{rep} \odot \mathbf{W}_{se} \in \mathbb{R}^{B \times C_i \times L_i}.
\end{equation}
This combination extracts local temporal details and highlights discriminative channels. Such refinement is critical for identifying subtle cardiac arrhythmias.

Finally, $\mathbf{X}_{se}$ is processed by the feed-forward network module $\mathcal{F}_{ffn}$ with a residual connection:
\begin{equation}
    \mathbf{X}_{even} = \mathcal{F}_{ffn}(\mathbf{X}_{se}) + \mathbf{X}_{se} \in \mathbb{R}^{B \times C_i \times L_i}.
\end{equation}
Notably, the first stage of the architecture includes only a single even-indexed block. This configuration establishes a robust local feature foundation before entering deeper alternating structures.

\subsubsection{Large-Kernel Perception}
\label{sssec:LKP}
The Large-Kernel Perception(LKP) module captures macroscopic structural patterns through an expansive receptive field to generate adaptive, signal-aware weights. This module derives dynamic weights from the temporal context of physiological signals. Given a feature map $\mathbf{X} \in \mathbb{R}^{B \times C_i \times L_i}$, the module maps the temporal context into dynamic weights $\mathbf{W}$ via a sequence of projections and a large-kernel depthwise convolution:
\begin{equation}
    \mathbf{X}_{lkp} = \text{BN-Conv}(\text{BN-DWConv}_{1\times7}(\text{BN-Conv}(\mathbf{X}))),
\end{equation}
\begin{equation}
    \mathbf{W}_{lkp} = \text{GroupNorm}(\text{Conv}(\mathbf{X}_{lkp})) \in \mathbb{R}^{B \times C'_i \times K_s \times L_i},
\end{equation}
where $C'_i = C_i/G$, $G$ is the number of groups, and $K_s=3$ represents the dynamic kernel size. Distilling macro-structural information into these time-varying weights enables the network to adapt to non-stationary variations in signal morphologies.

\subsubsection{Small-Kernel Aggregation}
\label{sssec:SKA}
The Small-Kernel Aggregation(SKA) module captures fine-grained morphological variations by performing dynamic weight aggregation. It applies the macroscopic insights from the LKP module to local, high-resolution temporal windows to implement dynamic convolution. For the input feature map $\mathbf{X} \in \mathbb{R}^{B \times C_i \times L_i}$, the signal is padded and unfolded into local windows $\mathbf{X}_{windows} \in \mathbb{R}^{B \times C_i \times L_i \times K_s}$. The dynamic aggregation is performed as follows:
\begin{equation}
\mathbf{X}_{ska}^{(b,c,l)} = \sum_{k=1}^{K_s} \mathbf{X}_{windows}^{(b,c,l,k)} \cdot \mathbf{W}_{lkp}^{(b, \lfloor c/G \rfloor, k, l)}.
\end{equation}

Finally, a residual connection and batch normalization are applied:
\begin{equation}
\mathbf{X}_{ls} = \text{BN}(\mathbf{X}_{ska}) \in \mathbb{R}^{B \times C_i \times L_i}.
\end{equation}
This localized aggregation filters redundant noise and emphasizes critical diagnostic features.

In the odd-indexed layers, the feature map $\mathbf{X}_{ls}$ passes through the feed-forward network $\mathcal{F}_{ffn}$ to stabilize the highly sensitive features. This output is processed as follows:
\begin{equation}
\mathbf{X}_{odd} = \mathcal{F}_{ffn}(\mathbf{X}_{ls})+\mathbf{X}_{ls} \in \mathbb{R}^{B \times C_i \times L_i}.
\end{equation}
This architecture maintains high sensitivity to minute cardiac anomalies.

\subsection{Integration with Transformer}
To model long-term heart rhythm dependencies, we map the encoder output $\mathbf{X}_{lsnet} \in \mathbb{R}^{B \times C_{final} \times L_{final}}$ into a Transformer architecture. A temporal positional encoding is applied, followed by $N$ Transformer blocks. Each layer utilizes a multi-head self-attention mechanism modified with LoRA:
\begin{equation}
    \mathbf{X}_{trans} = \text{Transformer}(\mathbf{X}_{lsnet}) + \mathbf{X}_{lsnet}.
\end{equation}
This integration bridges the fine-grained morphological features captured by the convolutional backbone with the global temporal context required for accurate arrhythmia detection.

\subsection{Parameter-Efficient Adaptation with LoRA}
Instead of updating the full parameter set of the pretrained backbone, we inject trainable low-rank matrices into the backbone-specific blocks, standard Transformer components, and classifier layers. In the convolutional backbone, all other convolutional and linear layers are enhanced with low-rank adapters, except for the initial patch embedding, the Squeeze-and-Excitation modules, and the core SKA operators. This design ensures that the backbone remains fixed to preserve robust pretrained morphologies while enabling task-specific adaptation. In the attention module, the query, key, and value projections are optimized using a merged adaptation strategy that allows simultaneous rank-adaptation of all three projections.

We apply a rank-allocation strategy guided by the intrinsic dimensionality of physiological representations\cite{ansuini2019intrinsic}. We assign a minimal rank of $r_c = 4$ to the convolutional layers. Because convolutional layers operate on local receptive fields to extract translation-invariant structural motifs such as QRS complexes and P and T waves, the intrinsic dimensionality of this localized structural manifold is low. Restricting $r_c$ to a minimal value acts as a regularizer that prevents the model from overfitting to high-frequency acquisition noise. Furthermore, by restricting the gradient updates to a low-rank subspace, LoRA effectively prevents the catastrophic forgetting of the robust, generalizable features learned during pretraining. This implicit regularization stabilizes the optimization trajectory and mitigates representation drift.

Conversely, a higher rank of $r_t = 16$ is granted to the Transformer-based mixers. These layers process global multi-head self-attention maps across the entire temporal domain, which exhibit high intrinsic dimensionality due to non-stationary rhythm variations and long-range beat dependencies. By focusing rank capacity on these global projections, we preserve the rich representational space of the global temporal dynamics without inflating the overall storage requirements. Ultimately, this dual-rank allocation leverages LoRA not only as a parameter-saving tool, but as a dual-purpose structural regularizer.

\subsection{Knowledge Distillation Framework}
To compress the high-capacity representation space into our lightweight student model while maintaining diagnostic sensitivity, we propose two distinct knowledge distillation paradigms, namely KD-Homo and KD-Hetero. 
Both strategies utilize a unified total loss function $\mathcal{L}$ as defined in Eq. \ref{eq:total_loss}. This function guides the student via a standard binary cross-entropy loss. This formulation effectively transfers the dark knowledge and nuanced probability distributions representing sub-clinical similarities from the teacher\cite{qiang2025df}.

The homogeneous paradigm, KD-Homo, employs a high-capacity teacher that shares the identical interleaved backbone structure as the student. This strategy enforces self-refinement on the same structural manifold. By distilling soft probabilities within aligned feature spaces, KD-Homo acts as a manifold smoothing regularizer, allowing the student to avoid local minima and out-perform fully supervised baselines.

Conversely, the heterogeneous paradigm, KD-Hetero, utilizes a heavy residual network and Transformer teacher to execute cross-architectural inductive bias transfer. While the residual backbone of the teacher enforces translation invariance through deep residual stacks, its subsequent Transformer encoder models global context. Distilling these heterogeneous features allows our lightweight student to inherit diverse, over-parameterized structural priors. The student effectively emulates the representational capacity of a deep residual hybrid model within its highly streamlined convolutional Transformer architecture.

\section{Experiment}
\label{experiment}
\subsection{Datasets and Preprocessing}
\label{subsec: datasets}
To initialize our model with robust physiological representations, we first pretrain on the large-scale CODE-15\% dataset \cite{ribeiro2021code}. To ensure a fair comparison, all baseline models are evaluated under equivalent large-scale pretraining regimes.
We validate our approach using three widely-used 12-lead ECG datasets, namely G12EC \cite{alday2020classification}, PTB-XL \cite{wagner2020ptb}, and Chapman-Shaoxing \cite{zheng2020chapman}. 
All dataset recordings are standardized to a sampling frequency of 500 Hz and cropped or padded to a uniform duration of 10 s. To ensure statistical significance and stable model training, we strictly include CVD classes that contain at least 200 samples.

Before data partitioning, the preprocessed ECG signals undergo a multi-stage processing pipeline. First, we calibrate the physical units using Analog-to-Digital Converter (ADC) gains and baselines. Next, we apply z-score normalization to each lead to standardize the signal distribution. 
For data-level augmentation, we incorporate random amplitude scaling and noise injection. Finally, during training, we employ a dynamic Cutmix strategy to improve model generalization.

\subsection{Experimental Setup and Evaluation Metrics}
\label{subsec: protocols and critrias}
All experiments are implemented in PyTorch using a fixed random seed of 42 for reproducibility. The evaluation relies on a patient-disjoint 10-fold cross-validation protocol to prevent any intra-subject information leakage across all ECG datasets. Within the development set of each fold, we allocate 20\% of the records to a validation subset for performance monitoring and early stopping. The remaining 80\% of the records are used for model training. We maintain this partitioning strategy consistently across all datasets to ensure a fair and robust comparison.

For optimization, we employ the AdamW optimizer alongside a linear warmup scheduler. The student models themselves are consistently trained using FT, resulting in identical training memory and latency footprints. During the teacher fine-tuning phase, the learning rate is 0.001, while during the subsequent distillation phase, the teacher is frozen and the student is trained with a learning rate of 0.002. Training utilizes a batch size of 64 and an early stopping patience of 30 epochs to achieve robust convergence. We conduct all experiments on an NVIDIA RTX 4090 GPU.

Average training runtime and peak GPU memory footprint serve as our decisive efficiency metrics. We use macro AUC and macro $F_{\beta=2}$ \cite{strodthoff2020deep} as primary performance metrics. Here, a beta value of two emphasizes recall to reduce false negatives in imbalanced clinical datasets. Additionally, we report secondary metrics to provide a comprehensive evaluation. These metrics include Hamming loss, which is denoted as HL, ranking loss, coverage, macro $G_{\beta=2}$, mean average precision, commonly known as MAP, and the total number of trainable parameters.

\subsection{Comparison with Baseline Methods}
\label{subsec: comparision}
We compare LSTrans against representative lightweight baselines. 
For specialized memory-efficient training, we evaluate H-Tuning\cite{zhou2025h} and AdaGrad-Fusion\cite{xu2026adagrad}. 
H-Tuning applies a mixed-order strategy combining first-order deep layer optimization, shallower mixed-order updates, LoRA, and knowledge distillation. 
AdaGrad-Fusion addresses the fine-tuning memory bottleneck by dynamically fusing zeroth-order search with selective backpropagation. 
We also compare against CE-SSL\cite{zhou2024ssl}, a distinct semi-supervised framework leveraging unlabeled ECG recordings via consistency regularization and pseudo-labeling. 
Finally, we evaluate against ECG-Founder\cite{li2024electrocardiogram}, a large-scale general-purpose foundation model. It is pretrained on over ten million expert-annotated clinical recordings from the Harvard-Emory database to capture comprehensive spatial and temporal lead representations.

\begin{table*}[t]
\centering
\caption{Comparison of LSTrans and Baseline Methods. Mem Denotes the Peak GPU Memory Footprint in GB. Time represents the average training runtime per Iteration in Seconds on an NVIDIA RTX 4090 GPU. Both Efficient Metrics are Reported as the Average Values Across the Three Datasets.}
\label{tab1:comparison1}
\setlength{\tabcolsep}{5pt}
\begin{tabular}{@{}l cc cc cc cc@{}}
\toprule
\multirow{2}{*}{\textbf{Model}} & 
\multicolumn{2}{c}{\textbf{G12EC}} & 
\multicolumn{2}{c}{\textbf{PTB-XL}} & 
\multicolumn{2}{c}{\textbf{Chapman-Shaoxing}} & 
\multirow{2}{*}{\textbf{Mem}} & 
\multirow{2}{*}{\textbf{Time}} \\
\cmidrule(lr){2-3} \cmidrule(lr){4-5} \cmidrule(lr){6-7}
& \textbf{AUC} & \textbf{$F_{\beta=2}$} & \textbf{AUC} & \textbf{$F_{\beta=2}$}  & \textbf{AUC} & \textbf{$F_{\beta=2}$} & & \\
\midrule
$\text{CE-SSL}_{r=16}$ \cite{zhou2024ssl} &0.855 $\pm$ 0.005 & 0.551 $\pm$ 0.017 & 0.901 $\pm$ 0.003 & 0.578 $\pm$ 0.006 & 0.896 $\pm$ 0.006 & 0.530 $\pm$ 0.008 & 2.7478 & 0.102\\
$\text{CE-SSL}_{r=4}$ \cite{zhou2024ssl} &0.853 $\pm$ 0.004 & 0.553 $\pm$ 0.020 & 0.899 $\pm$ 0.004 & 0.580 $\pm$ 0.006 & 0.898 $\pm$ 0.005 & 0.530 $\pm$ 0.012 & 2.743 & 0.101\\
H-Tuning \cite{zhou2025h} & 0.870 $\pm$ 0.002 & 0.586 $\pm$ 0.010 & 0.923 $\pm$ 0.003 & 0.628 $\pm$ 0.002 & 0.929 $\pm$ 0.002 & 0.634 $\pm$ 0.009 & 1.453 & 0.401\\
AdaGrad-Fusion \cite{xu2026adagrad} & 0.874 & 0.594 & 0.923 & 0.633 & 0.936 & 0.649 & 0.787 & 0.173\\
ECG-Founder \cite{li2024electrocardiogram} & 0.901 & 0.627 & 0.948 & 0.696 & 0.962 & 0.740 & 1.347 & 0.250\\
\midrule
ResNet-Teacher (FT) & 0.906 $\pm$ 0.007 & 0.664 $\pm$ 0.007 & 0.949 $\pm$ 0.004 & 0.702 $\pm$ 0.009 & 0.960 $\pm$ 0.005 & 0.739 $\pm$ 0.019 & 1.571 & 0.177\\
ResNet-Teacher (LoRA) & 0.913 $\pm$ 0.003 & 0.673 $\pm$ 0.013 & 0.954 $\pm$ 0.003 & 0.711 $\pm$ 0.010 & 0.961 $\pm$ 0.005 & 0.736 $\pm$ 0.019 & 1.528 & 0.177\\
LSNet-Teacher (LoRA) & 0.868 $\pm$ 0.012 & 0.588 $\pm$ 0.016 & 0.925 $\pm$ 0.009 & 0.640 $\pm$ 0.015 & 0.930 $\pm$ 0.007 & 0.645 $\pm$ 0.016 & 1.342 & 0.141\\
\midrule
LSTrans(Hetero+FT) & 0.903 $\pm$ 0.010 & 0.655 $\pm$ 0.013 & 0.950 $\pm$ 0.004 & 0.703 $\pm$ 0.008 & 0.955 $\pm$ 0.006 & 0.720 $\pm$ 0.018 & 0.856 & 0.084\\
LSTrans(Hetero+LoRA) & 0.928 $\pm$ 0.010 & 0.696 $\pm$ 0.019 & 0.956 $\pm$ 0.002 & 0.713 $\pm$ 0.008 & 0.963 $\pm$ 0.005 & 0.740 $\pm$ 0.017 & 0.856 & 0.084\\
LSTrans(Homo+LoRA) & 0.880 $\pm$ 0.009 & 0.612 $\pm$ 0.018 & 0.930 $\pm$ 0.013 & 0.656 $\pm$ 0.021 & 0.943 $\pm$ 0.009 & 0.687 $\pm$ 0.014 & 0.836 & 0.071\\
\bottomrule
\end{tabular}
\end{table*}

Experimental results in Tab. \ref{tab1:comparison1} indicate that LSTrans hetero achieves highly competitive performance. Specifically, LSTrans hetero achieves an average AUC of 0.949 and an average $F_{\beta=2}$ score of 0.716, often matching or exceeding established baselines.
For comparison, LSTrans hetero outperforms AdaGrad-Fusion and $\text{CE-SSL}_{r=16}$ in terms of average AUC by 0.038 and 0.065, respectively. More notably, the average $F_{\beta=2}$ score of LSTrans hetero represents relative improvements of approximately 14.6\% and 29.5\% over these two baselines.

Compared to H-Tuning, LSTrans hetero demonstrates clear performance advantages, particularly in terms of the primary clinical safety metric $F_{\beta=2}$. Specifically, LSTrans hetero outperforms H-Tuning by absolute margins of 11.0\% on the G12EC dataset, 8.5\% on the PTB-XL dataset, and 10.6\% on the Chapman-Shaoxing dataset. This improvement reflects the capacity of our model to recognize fine-grained wave anomalies, which helps minimize false-negative diagnostic errors. Regarding macro $G_{\beta=2}$, LSTrans hetero yields an average relative gain of 7.8\% across the three datasets. LSTrans hetero similarly elevates the MAP across all three datasets, highlighted by a notable 9.1\% absolute MAP increase on the G12EC dataset.

Furthermore, we evaluate LSTrans hetero against ECG-Founder. Despite its significantly smaller parameter scale, LSTrans hetero consistently matches the performance of this foundation model. A closer inspection of fine-grained secondary metrics reveals that LSTrans hetero closely approaches the baseline performance established by this heavy foundation model. For the PTB-XL dataset, our lightweight LSTrans hetero achieves a highly competitive macro $G_{\beta=2}$ score, narrowing the absolute gap to only 0.036 points. At the same time, it maintains comparable error rates and coverage limits. A similar trend is observed on the Chapman-Shaoxing dataset, where LSTrans hetero trails the foundation model by a margin of only 13.7\%.
Crucially, this competitive diagnostic performance is achieved with a drastically minimized model footprint. While ECG-Founder scales up to 30.68M parameters, LSTrans hetero utilizes a much lighter configuration. Specifically, our framework incorporates a teacher model of 3.49M parameters and an even more streamlined student model of 3.19M parameters. Furthermore, compared to the ECG-Founder baseline, our framework achieves a peak memory reduction of approximately 36.4\% and speeds up execution by nearly 3 times.

Beyond diagnostic accuracy, LSTrans hetero achieves a significantly lower computational overhead and faster adaptation speed during downstream task. When compared to the specialized lightweight framework H-Tuning, LSTrans hetero reduces the peak GPU memory footprint by approximately 41.09\% relative to H-Tuning's 1.453 GB. Additionally, LSTrans hetero accelerates the average runtime by nearly 4.77 times. In comparison with AdaGrad-Fusion, which is specifically optimized to minimize training memory, LSTrans hetero achieves a comparable peak memory footprint while delivering a significant speedup of 2.06 times.

\subsection{Ablation Studies}
\begin{figure}
\centering
\includegraphics[width=0.5\textwidth]{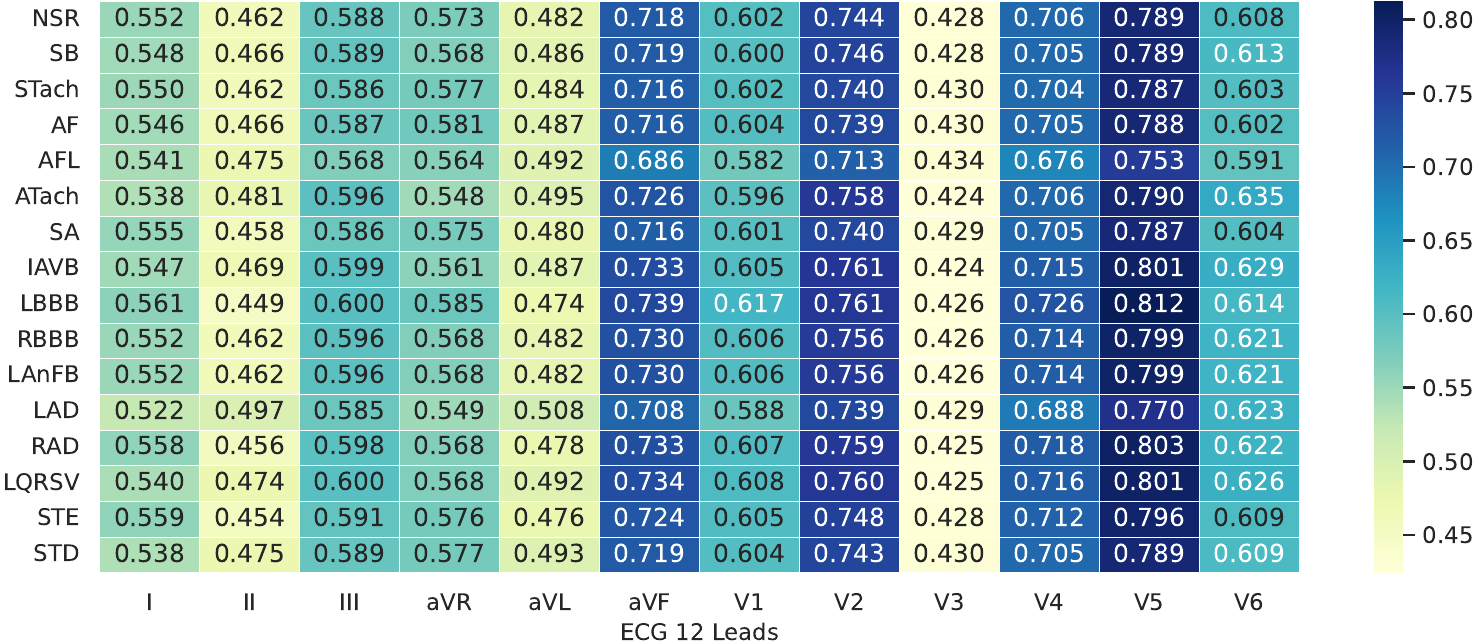} 
\caption{SE Gate Activation Weights}
\label{fig:SE Gate Activation Weights}
\end{figure}

In the even-indexed layers, removing RepVGGDW evaluates local multi-scale features extraction. Without RepVGG, $F_{\beta=2}$ decreases by 6.61\% to 0.650. Reparameterized convolutions are vital for capturing sharp wave complexes. Omitting Squeeze-and-Excitation evaluates lead recalibration. This omission lowers $F_{\beta=2}$ score by 6.75\% to 0.649. Fig. \ref{fig:SE Gate Activation Weights} validate that this module helps suppress background artifacts. 
Replacing all even-indexed layers with the odd-indexed layer structure yields a homogeneous network and degrades $F_{\beta=2}$ score by 7.61\% to 0.643, confirming the necessity of even layers to prevent representation over-smoothing.

\begin{figure}
\centering
\includegraphics[width=0.5\textwidth]{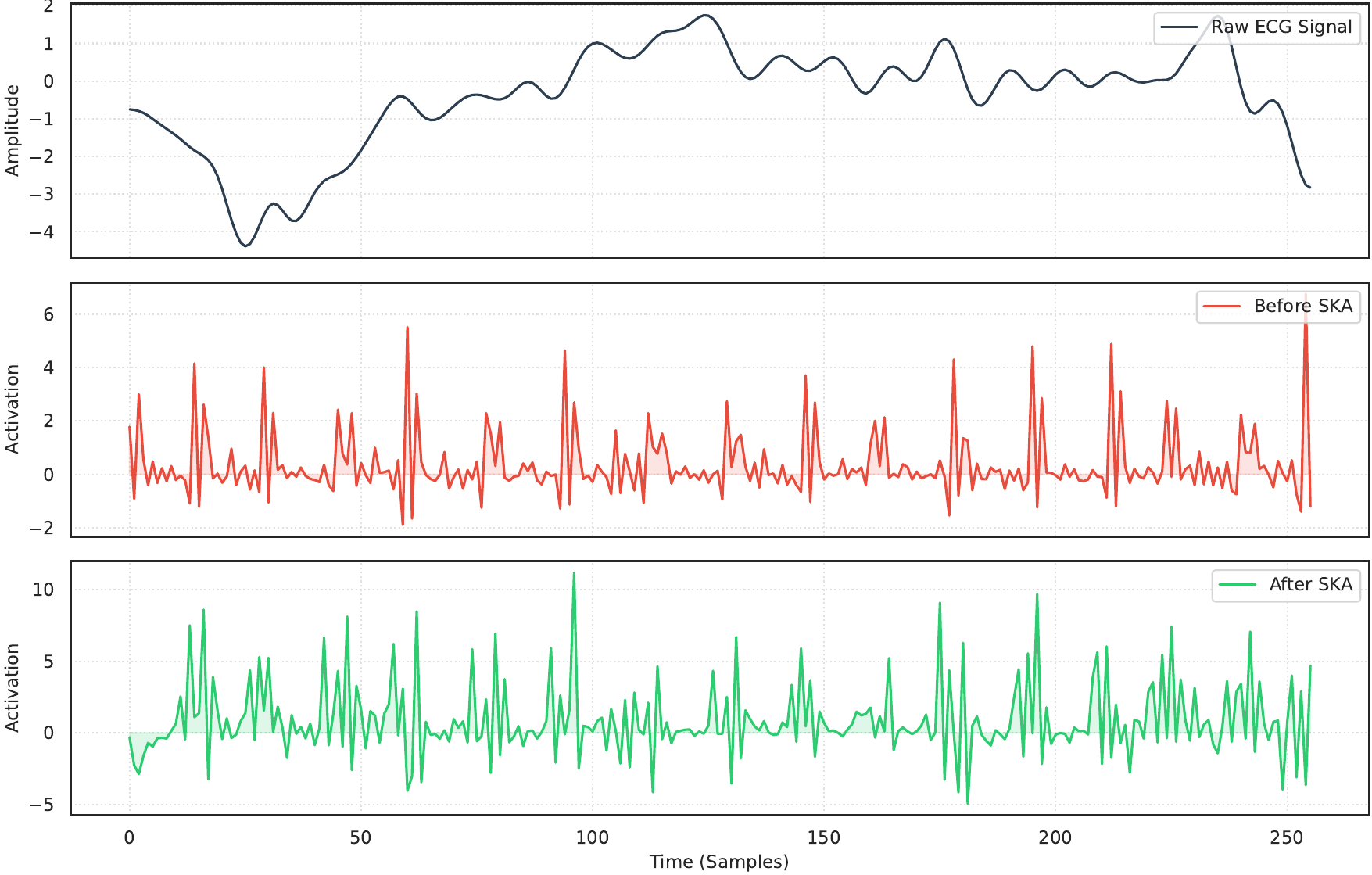} 
\caption{SKA Feature Enhancement Analysis}
\label{fig:SKA Feature Enhancement Analysis}
\end{figure}

\begin{figure}
\centering
\includegraphics[width=0.5\textwidth]{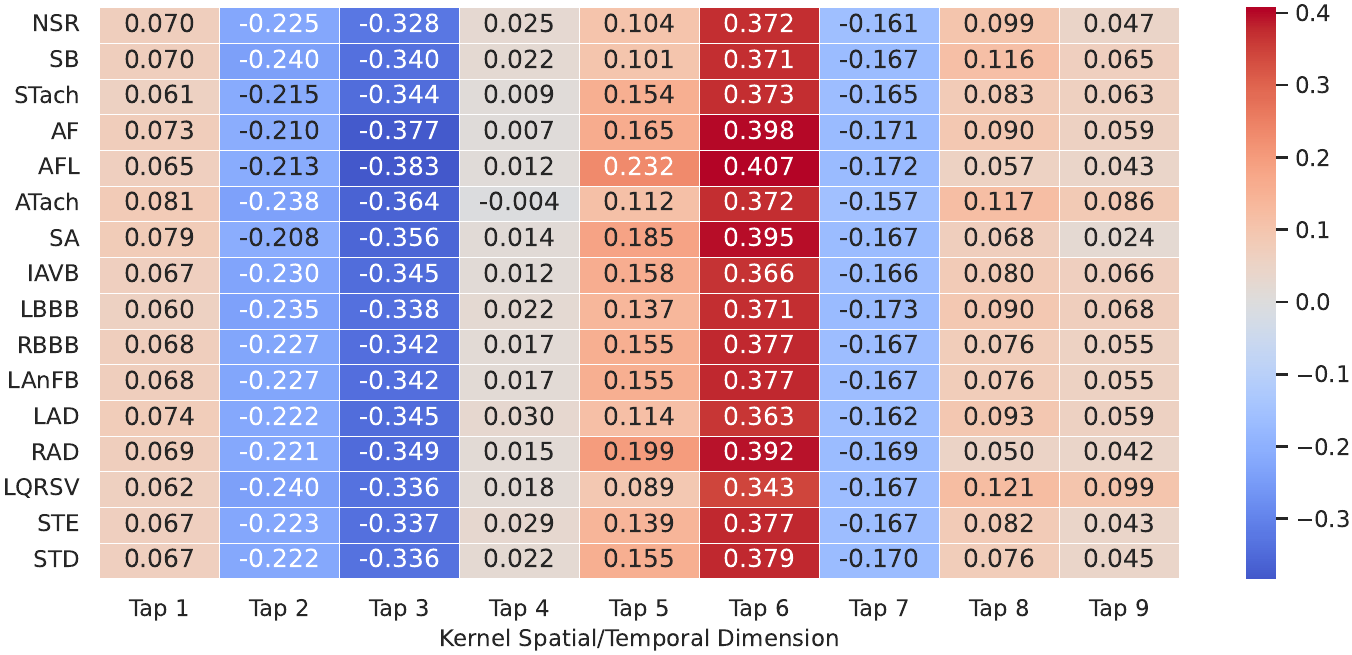} 
\caption{LKP Kernel Weights}
\label{fig:LKP Kernel Weights}
\end{figure}

In the odd-indexed layers, removing SKA degrades the AUC to 0.905 and decreases $F_{\beta=2}$ score by 6.18\% to 0.653. This decline occurs because SKA is essential for peak preservation and noise suppression, as visualized in Fig. \ref{fig:SKA Feature Enhancement Analysis}. 
Replacing LKP with static convolutions tests the global context bottleneck. This reduces the AUC by 2.48\% and lowers $F_{\beta=2}$ by 5.89\% to 0.655.
The adaptivity of LKP is demonstrated in Fig. \ref{fig:LKP Kernel Weights}, where weights dynamically adjust to capture non-stationary rhythms.
Replacing the entire LSConv structure with standard 3 $\times$ 3 convolutions drops the AUC by 2.69\% and decreases $F_{\beta=2}$ by 6.47\%.

As an illustrative example, Left Bundle Branch Block (LBBB) clinically manifests as deep S-waves in septal leads $V_1$ and $V_2$, alongside broad R-waves in lateral leads $V_5$ and $V_6$. The model assigns prominent attention weights $0.812$ to lead $V_5$ and $0.761$ to lead $V_2$ under the LBBB category.
As shown in Fig. \ref{fig:SKA Feature Enhancement Analysis}, the SKA module suppresses chaotic, high-frequency baseline noise and concentrates activation energy on the diagnostic QRS peak.

Beyond diagnostic performance, the ultimate reduction in computational overhead is a combined effect of these odd- and even-layer components. Individually, replacing the LSConv increases peak memory by 1.64\% and iteration latency by 3.57\%. More significantly, replacing even-layer structures expands peak memory by 23.61\% and slows execution by 19.05\%.
These compounding overheads demonstrate that our overall efficiency is a synergistic outcome.

\begin{table}[t]
\centering
\caption{Ablation Study on LSNet in G12EC Dataset.}
\label{tab3:ablation}
\resizebox{\columnwidth}{!}{%
\begin{tabular}{@{}l cc cc@{}}
\toprule
\textbf{Model} & \textbf{AUC} & \textbf{$F_{\beta=2}$} & \textbf{Mem} & \textbf{Time} \\
\midrule
w/o SKA & 0.905 $\pm$ 0.000 & 0.653 $\pm$ 0.000 & 0.569 & 0.079 \\
w/o LKP & 0.905 $\pm$ 0.001 & 0.655 $\pm$ 0.005 & 0.537 & 0.076 \\
w/o LSConv & 0.903 $\pm$ 0.002 & 0.651 $\pm$ 0.010 & 0.680 & 0.087 \\
\midrule
w/o RepVGG & 0.906 $\pm$ 0.002 & 0.650 $\pm$ 0.002 & 0.656 & 0.083 \\
w/o se & 0.902 $\pm$ 0.002 & 0.649 $\pm$ 0.007 & 0.655 & 0.083 \\
w/o even-indexed layers & 0.902 $\pm$ 0.001 & 0.643 $\pm$ 0.001 & 0.827 & 0.100 \\
\midrule
LSTrans (Hetero+LoRA) & 0.928 $\pm$ 0.010 & 0.696 $\pm$ 0.019 & 0.669 & 0.084 \\
\bottomrule
\end{tabular}
}
\end{table}

\subsection{Impact of Knowledge Distillation and LoRA}
For performance improvements, the choice of distillation strategy affects student representation learning. The LSTrans hetero utilizes cross-architectural inductive bias transfer. This approach achieves relative improvements of 3.41\% in mean AUC and 9.91\% in mean $F_{\beta=2}$ over the LSTrans homo. This gain indicates that transferring over-parameterized structural priors from a deep residual hybrid teacher restores diagnostic sensitivity for rare cardiac anomalies.

Next, the injection of trainable LoRA adapters yields classification advantages compared to full fine-tuning. Specifically, LSTrans hetero distilled from the LoRA-adapted teacher achieves relative improvements of 1.39\% in mean AUC and 3.41\% in mean $F_{\beta=2}$ over its fully fine-tuned counterpart LSTrans. This gap suggests that restricting gradient updates to a low-rank subspace acts as an implicit, dual-purpose structural regularizer. This restriction prevents catastrophic forgetting and stabilizes the optimization trajectory against high-frequency acquisition noise.

For computational overhead, transferring knowledge to the lightweight student configuration reduces the peak GPU memory footprint by 43.98\% compared to the ResNet-Teacher. It also accelerates the average training runtime by 52.5\%. Under restricted training resources, the LSNet-Teacher slashes the memory footprint by 12.17\% and accelerates training by 20.34\% compared to the ResNet-Teacher. 
This alternative offers a flexible trade-off between deployment complexity and optimization efficiency.

\section{Conclusion}
\label{sec: con}
This paper presents LSTrans, a lightweight hybrid framework for automated ECG classification. By utilizing an interleaved 1D-LSNet backbone cascaded with a Transformer encoder, our approach effectively balances diagnostic accuracy with adaptation efficiency.
Extensive evaluations across multiple clinical benchmarks demonstrate that LSTrans achieves highly competitive diagnostic performance compared to heavy baseline models, while significantly reducing parameter counts and fine-tuning footprints.

Several key limitations remain for future work. First, our efficiency evaluations rely on training-phase metrics, leaving actual inference latency on non-GPU edge hardware unmeasured. Second, runtime weight generation in dynamic convolutions may incur control-flow and memory overhead on resource-constrained platforms. Finally, generalization remains dependent on large-scale pretraining. Future efforts will focus on optimizing these dynamic operators, measuring real-world edge latencies, and conducting clinical validation on wearable ECG patches.

\balance
\bibliographystyle{IEEEtran}
\bibliography{LSTrans}

\end{document}